\documentclass[journal,twoside,web]{ieeecolor}
\usepackage{tmi}
\usepackage{cite}
\usepackage{amsmath,amssymb,amsfonts}
\usepackage{algorithmic}
\usepackage{graphicx}
\usepackage{textcomp}
\usepackage[utf8]{inputenc} 
\usepackage[T1]{fontenc}    
\usepackage{booktabs}       
\usepackage{amsfonts}       
\usepackage{nicefrac}       
\usepackage{microtype}      
\usepackage{colortbl,booktabs}
\usepackage{environ}
\usepackage{multicol}
\usepackage{multirow}
\usepackage{graphicx} 
\usepackage{tikz}
\usepackage{wrapfig}
\usepackage{makecell}
\usepackage{diagbox}
\usepackage{pifont}
\usepackage{diagbox} 
\usepackage{tabularx}
\usepackage{wrapfig,lipsum,booktabs}
\usepackage{comment}
\usepackage{xcolor}
\usepackage{hyperref}

\usepackage{etoolbox}
\makeatletter
\@ifundefined{color@begingroup}%
  {\let\color@begingroup\relax
   \let\color@endgroup\relax}{}%
\def\fix@ieeecolor@hbox#1{%
  \hbox{\color@begingroup#1\color@endgroup}}
\patchcmd\@makecaption{\hbox}{\fix@ieeecolor@hbox}{}{\FAILED}
\patchcmd\@makecaption{\hbox}{\fix@ieeecolor@hbox}{}{\FAILED}

\def\BibTeX{{\rm B\kern-.05em{\sc i\kern-.025em b}\kern-.08em
    T\kern-.1667em\lower.7ex\hbox{E}\kern-.125emX}}
\begin{document}
\title{Radiomics-Guided Global-Local Transformer for Weakly Supervised Pathology Localization in Chest X-Rays}
\author{Yan Han, Gregory Holste, Ying Ding, Ahmed Tewfik, \textit{Fellow, IEEE}, \\ Yifan Peng, and Zhangyang Wang, \textit{Senior Member, IEEE}
\thanks{Manuscript submitted <date>.}
\thanks{Yan Han, Gregory Holste, Ying Ding, Ahmed Tewfik, and Zhangyang Wang are with The University of Texas at Austin, Austin, TX 78705 USA (email: yh9442@utexas.edu; gholste@utexas.edu; ying.ding@ischool.utexas.edu; tewfik@austin.utexas.edu; atlaswang@utexas.edu).}
\thanks{Yifan Peng is with Weill Cornell Medical College, New York, NY 10065 (email: yip4002@med.cornell.edu).}
}

\maketitle

\begin{abstract}
Before the recent success of deep learning methods for automated medical image analysis, practitioners used handcrafted radiomic features to quantitatively describe local patches of medical images. However, extracting discriminative radiomic features relies on accurate pathology localization, which is difficult to acquire in real-world settings. Despite advances in disease classification and localization from chest X-rays, many approaches fail to incorporate clinically-informed domain-specific radiomic features. For these reasons, we propose a Radiomics-Guided Transformer (RGT) that fuses \textit{global} image information with \textit{local} radiomics-guided auxiliary information to provide accurate cardiopulmonary pathology localization and classification \textit{without any bounding box annotations}. RGT consists of an image Transformer branch, a radiomics Transformer branch, and fusion layers that aggregate image and radiomics information. Using the learned self-attention of its image branch, RGT extracts a bounding box for which to compute radiomic features, which are further processed by the radiomics branch; learned image and radiomic features are then fused and mutually interact via cross-attention layers. Thus, RGT utilizes a novel end-to-end feedback loop that can bootstrap accurate pathology localization only using image-level disease labels. Experiments on the NIH ChestXRay dataset demonstrate that RGT outperforms prior works in weakly supervised disease localization (by an average margin of 3.6\% over various intersection-over-union thresholds) and classification (by 1.1\% in average area under the receiver operating characteristic curve). We publicly release our codes and pre-trained models at \url{https://github.com/VITA-Group/chext}.
\end{abstract}


\section{Introduction}
In medicine, \textit{radiomics} refers to the process of extracting quantitative and semiquantitative features from medical images, such as radiographs or computed tomography scans, for improved decision support \cite{zwanenburg2016image}.
These handcrafted radiomic features aim to describe a local ``region of interest'' such as a tumor with numeric features that assess qualities such as size, shape, texture, variations in pixel intensity, and relationships between neighboring pixels \cite{parekh2016radiomics}. Given their advantages, researchers have explored the performance of radiomic features for chest X-ray analysis. For example, Shi \textit{et al.} \cite{shilarge} and Sayg{\i}l{\i} \cite{saygili2021new} each extracted a set of radiomic features, which were then used to diagnose different types of pneumonia. Bai \textit{et al.} \cite{bai2020predicting} proposed a hybrid model to encode the combination of radiomic features and clinical information. Ghosh \textit{et al.} \cite{ghosh2020quantitative} presented a new handcrafted feature to distinguish between severe and nonsevere patients. However, all of the above methods rely on \textit{accurate pathology localization annotations} to extract radiomic features from a correct and clinically meaningful region of interest \cite{van2017computational}. Such bounding boxes are usually expensive and time-consuming to acquire by humans and, if inaccurate, will tremendously degrade the reliability of radiomic features. There is thus an unmet need to automatically localize cardiopulmonary pathologies on chest X-rays to facilitate extraction of radiomic features.

Throughout the rapid development of deep learning approaches for medical image analysis, many researchers have made efforts utilizing convolutional neural networks (CNNs) to build automated systems for chest X-ray abnormality classification and localization \cite{rajpurkar2017chexnet, wang2017chestx, li2018thoracic, liu2019align, rozenberg2020localization, wang2021knowledge, yu2022anatomy, chandra2022disease, fernando2022chest, kumarasinghe2022u, meedeniya2022chest}. However, CNN methods bear several limitations when applied to the domain of chest radiography.
First, CNNs do not naturally incorporate contextual prior information, such as reason for imaging and patient history, or domain knowledge such as human anatomy and typical disease presentation on imaging. Since radiomic features are designed by humans and semantically describe local medical image regions, they represent an auxiliary modality of information embedded with domain-specific quantitative features that can enhance automated disease localization and classification. Second, chest X-rays have more subtle discriminative features compared to natural images, making their recognition more challenging.
Finally, though many have studied the interpretability of deep image classifiers for other data \cite{jing2019coarse,ribeiro2016should,ribeiro2018anchors,montavon2017explaining,liu2019generative,lundberg2017unified}, deep CNNs are often criticized for their lack of human interpretability, thus posing a major barrier to their adoption by clinicians.

With this in mind, Transformers, which have seen a surge in popularity for a variety of visual recognition tasks, provide a promising alternative to CNNs for modeling chest X-rays. The Transformer was first introduced in the context of natural language processing \cite{vaswani2017attention, devlin2018bert, brown2020language}, followed by its recent success in computer vision \cite{dosovitskiy2020image, carion2020end, zhu2020deformable} and multi-modal learning \cite{ying2021transformers}. The Transformer architecture can be considered a ``universal modeling tool'' that can unify the feature extraction and fusion processes from different input modalities with a \textit{single} model that does not require domain-specific architecture tweaks. For example, Arkbari \textit{et al.} \cite{akbari2021vatt} demonstrated the ability to learn powerful multi-modal representations from unlabeled video, audio, and text data, using a single multimodal Transformer. Nagrani \textit{et al.} design a bottleneck fusion technique that allows audio- and video-derived features to interact throughout their custom Transformer architecture \cite{nagrani2021attention}. And Shvetsova \textit{et al.} \cite{shvetsova2021everything} proposed a multi-modal, modality agnostic fusion Transformer to learn to exchange information between multiple modalities, such as video, audio, and text, and integrate them into a jointly multi-modal representation to obtain an embedding that aggregates multi-modal temporal information.

In the context of modeling chest X-rays, we observe the unique potential for a Transformer-based architecture to \textit{naturally and jointly learn from two ``views'' of chest X-rays}: (1) raw X-ray images that contain rich contrast details, hence benefiting from the data-driven learning capacity, and (2) radiomics that encode domain-specific quantitative features, thus guiding and regularizing the learning process with handcrafted local radiomic features. However, there exists a ``chicken-and-egg'' problem: extraction of useful radiomic features relies on accurate pathology localization, but the pathology localization is often absent and first needs to be learned or separately acquired.

\begin{figure}
    \centering
    \includegraphics[width=.99\linewidth,page=1,clip,trim=3cm 3cm 16cm 5em]{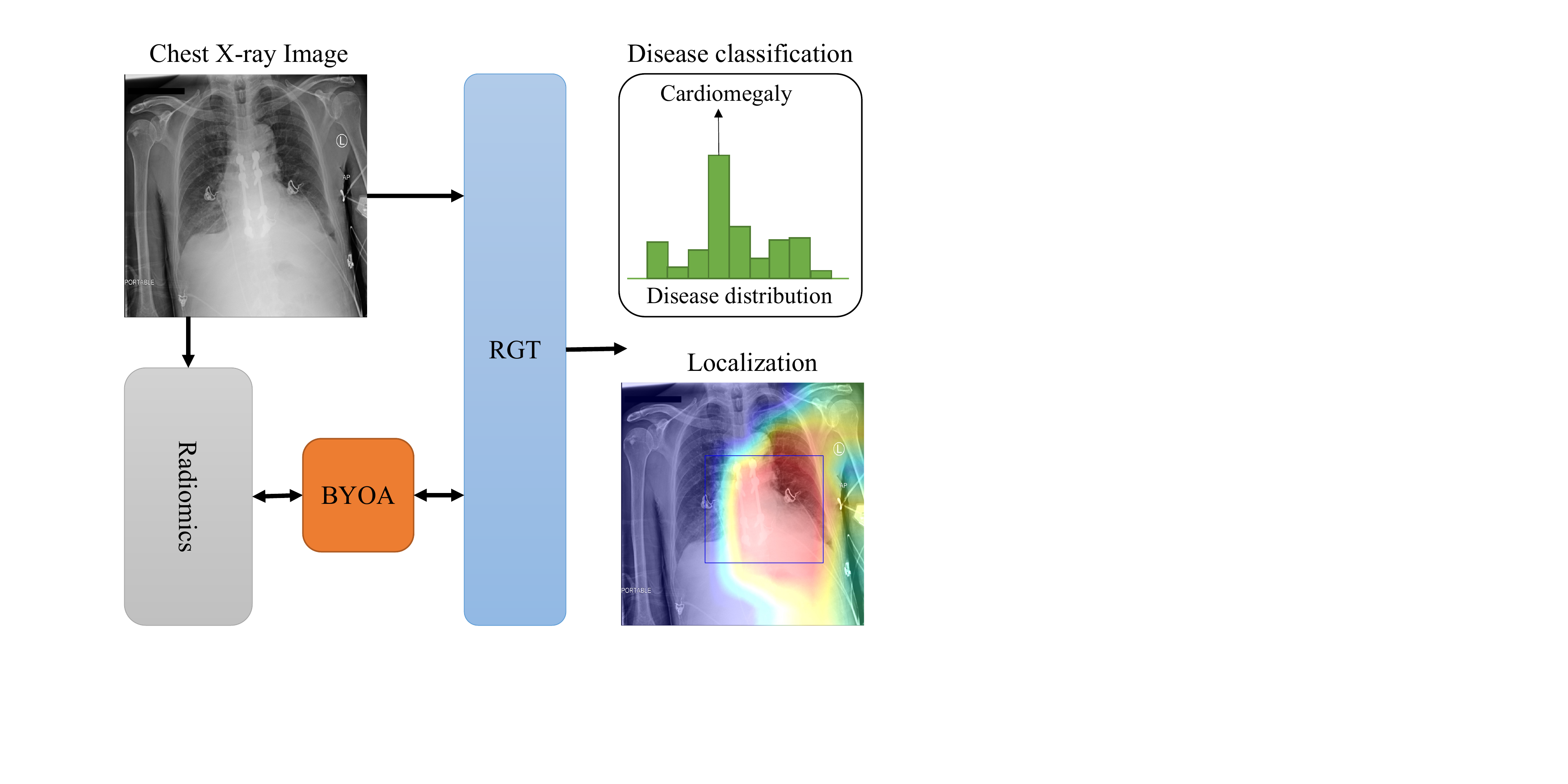}
    \caption{General overview of our \textbf{R}adiomics-\textbf{G}uided \textbf{T}ransformer (\textbf{RGT}) framework for weakly supervised cardiopulmonary disease localization and classification from chest X-rays. RGT takes a chest X-ray as the input and produces a heatmap for pathology localization, from which a bounding box is obtained. Radiomic features are further extracted from the bounded region and fused with image-derived features to classify the pathology present. The detailed views of \textbf{RGT} framework and \textit{\textbf{B}ring \textbf{Y}our \textbf{O}wn \textbf{A}ttention} (\textbf{BYOA}) module are given in Fig.~\ref{rgt} and Fig.~\ref{byoa}, respectively.
    }
    \label{overview}
\end{figure}

This work presents \textbf{RGT}, a \textbf{R}adiomics-\textbf{G}uided \textbf{T}ransformer (Fig.~\ref{overview}). RGT consists of two Transformer-based branches, one for the raw chest X-ray and one for the radiomic features extracted from the corresponding image. Features extracted from these two ``views'' of the patient are then deeply fused with interaction via cross-attention layers \cite{chen2021crossvit}. Of note, the radiomic features need to be extracted from the learned pathology localizations, which are not readily available. The \textit{key enabling technology} to resolve this hurdle is to construct a feedback loop, called the \textit{\textbf{B}ring \textbf{Y}our \textbf{O}wn \textbf{A}ttention} (\textbf{BYOA}) module, which will be expanded in Sec~\ref{model}. During training, the image branch leverages its learned self-attention to estimate pathology location, which is then used to extract radiomic features from the original image for further processing by the radiomics branch. In addition to a supervised classification loss, we optimize the model with a contrastive loss that rectifies the image-derived and radiomics-derived ``views'' of the patient, and such an end-to-end optimization loop can bootstrap accurate pathology localizations from image data with \textit{no bounding box annotations} used for training.


Our contributions are outlined as follows:
\begin{itemize}
    \item We leverage radiomics as an ``auxiliary input modality'' that both correlates with the raw image modality and encodes domain-specific quantitative features. We then propose a novel radiomics-guided cross-attention Transformer, \textbf{RGT}, to jointly extract and fuse global image features and local radiomic features for disease localization and classification in chest X-rays.
    \item To resolve the key ``chicken-and-egg'' problem of extracting radiomic features without available cardiopulmonary pathology localization, we construct an innovative optimization loop where the learned image-level attention map is used to extract local radiomic features. Such an end-to-end loop can bootstrap accurate cardiopulmonary pathology localization from images without leveraging human-annotated bounding boxes.
    \item On the NIH ChestXRay benchmark \cite{wang2017chestx}, our approach achieves superior disease localization and classification results. RGT outperforms prior work in weakly supervised localization by an average margin of 3.6\% over different intersection-over-union (IoU) thresholds.
\end{itemize}

\begin{figure*}[t!]
\centering
\includegraphics[width=\linewidth,page=2,clip,trim=1cm 7cm 0cm 0em]{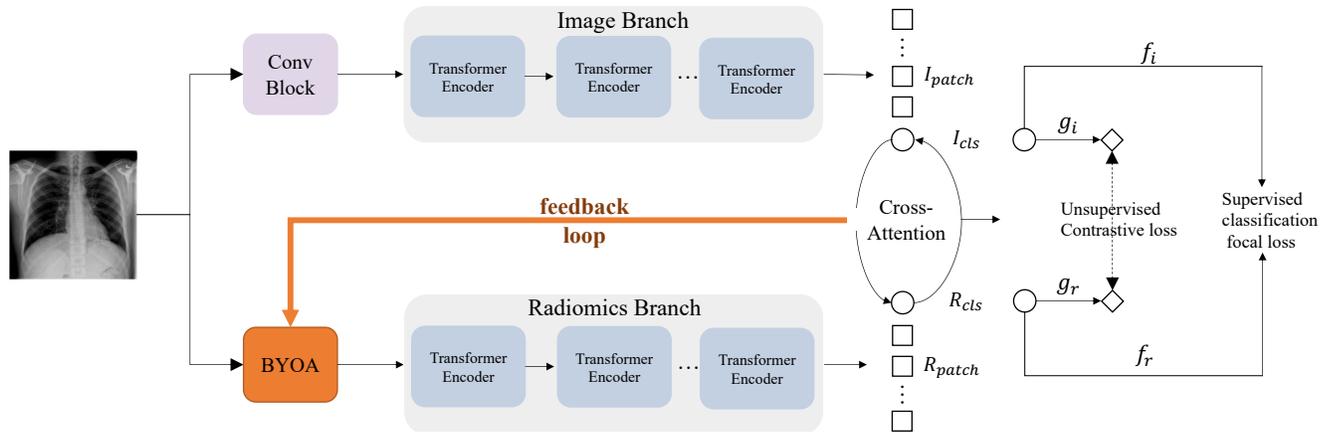}
\caption{Overview of our proposed model, RGT. The image branch is a Transformer that processes a chest X-ray, and the radiomics branch is a small Transformer that processes radiomic features generated by the Bootstrap Your Own Attention (BYOA) module (Fig.~\ref{byoa}). The global image representations and local radiomics representations are then fused by an efficient cross-attention module operation on each branch's \texttt{CLS} tokens. Finally, the \texttt{CLS} tokens $I_{cls}$ (from the image branch) and $R_{cls}$ (from the radiomics branch) are used for disease classification. We optimize the classification error with the Focal Loss \cite{lin2017focal}. We also leverage a contrastive learning strategy that aims to rectify the global image view with the local radiomics view. Specifically, RGT generates an image view $z_i = g_i(I_{cls})$ by a projection head $g_i$ and radiomic view $z_r = g_r(R_{cls})$ by projection head $g_r$. We maximize the agreement between $z_i$ and $z_r$ via a contrastive loss (NT-Xent).}
\label{rgt} 
\end{figure*}

\section{Related Work}


\textbf{Radiomics in Medical Diagnosis.} The design of radiomic features involves prior biological and medical knowledge, thus enriching the value provided by the raw pixel intensities of a medical image {\cite{gillies2016radiomics}}. In the study of image-based biomarkers for cancer staging and prognostication, radiomics has shown promising predictive power {\cite{nasief2019machine}}. Radiomics extracts quantitative features from medical images that can be used to represent tumor phenotypes, such as spatial heterogeneity of the tumor and spatial response variations. Eilaghi {\textit{et al.} \cite{eilaghi2017ct}} demonstrated that radiomic texture features, extracted from computed tomography (CT) scans, are associated with overall survival rate of pancreatic cancer. Chen {\textit{et al.} \cite{chen2017assessment}} revealed that the first-order radiomic features (e.g., mean, skewness, and kurtosis) are correlated with pathological responses to cancer treatment. Huang {\textit{et al.} \cite{huang2018added}} showed that radiomics could increase the positive predictive value and reduce the false-positive rate in lung cancer screening for small nodules compared to radiologists.  Zhang {\textit{et al.} \cite{zhang2018learning}} found that radiomics from multiparametric magnetic resonance imaging-based nomograms provided improved prognostic ability in advanced nasopharyngeal carcinoma. In positron emission tomography (PET)/CT imaging, Alongi {\textit{et al.} \cite{alongi2021radiomics}} leveraged radiomic features to successfully predict prostate cancer progression.


In comparison, deep learning is often criticized for being a ``black box'', lacking interpretability or reasoning in the form of human semantic concepts, despite achieving high predictive performance. This limitation has motivated many interpretable learning techniques including activation maximization \cite{erhan2009visualizing}, network inversion \cite{mahendran2015understanding}, GradCAM \cite{selvaraju2016grad}, and network dissection \cite{bau2017network}. We believe that the joint utilization of radiomics and interpretable learning techniques in our framework can further advance accurate yet interpretable learning in the medical image domain.

\textbf{Transformers for Medical Images.}
Recently, the Vision Transformer (ViT) \cite{dosovitskiy2020image} achieved state-of-the-art classification on ImageNet  by directly applying Transformers with global self-attention to full-sized images. Before the advent of Transformers for visual recognition tasks, several works have augmented traditional CNNs with attention modules, seeing improved performance on various medical image analysis problems \cite{oktay2018attention, wang2019volumetric}.
Now inspired by the promising performance of ViT, researchers have begun to adapt the Transformer architecture to medical image analysis problems.
For example, Chen \textit{et al.} \cite{chen2021transunet} and Hatamizadeh \textit{et al.} \cite{hatamizadeh2022unetr} demonstrate that Transformers can achieve state-of-the-art medical image segmentation performance over CNN-based architectures.
Similarly, Valanarasu \textit{et al.} \cite{valanarasu2021medical} proposed a gated axial-attention Transformer model to introduce an additional control mechanism in the self-attention module. Beyong medical image segmentation, Park \textit{et al.} \cite{park2021vision} utilized a hybrid CNN and Transformer framework for COVID-19 prediction. While these methods see improved performance by leveraging the Transformer architecture, none of these approaches incorporate domain-specific radiomic features. Han \textit{et al.} \cite{han2021pneumonia} applied pre-extracted radiomic features to guide pneumonia detection from chest X-ray images. However, they adopted a convolutional backbone for image encoder,  while using a specifically crafted radiomics encoder. Therefore, the method involves no joint interaction between image and radiomic features, requiring the use of previously acquired bounding boxes during training in order to extract radiomic features, limiting the method's usability in clinical practice. 

\section{Method}

An overview of RGT is illustrated in Fig.~\ref{rgt}. In the following subsections, we will first present Cross-Attention Vision Transformer (CrossViT), a recent two-branch ViT backbone on which RGT is built, and then describe
the methodological innovations required to naturally incorporate domain-specific quantitative features in the form of radiomics for improved cardiopulmonary pathology localization and classification.

\subsection{Preliminary: ViT and Cross-Attention}

ViT first converts an image into a sequence of patch tokens by dividing the image into fixed-size patches and linearly projecting each patch into so-called ``tokens''. A special \texttt{CLS} (class) token is prepended to the sequence of image patches, as in the original BERT \cite{devlin2018bert}. Then, all tokens are passed through stacked Transformer encoder layers. Finally, the hidden state corresponding to the \texttt{CLS} token is used as the aggregate sequence representation used for image classification. 

A Transformer encoder is composed of a sequence of blocks, where each block consists of (1) a multi-headed self-attention and (2) a feed-forward neural network. Layer normalization and residual shortcuts are, respectively, applied before and after every block. The granularity of the patch size affects the accuracy and complexity of ViT. Therefore, ViT was observed to reach greater performance with smaller (more fine-grained) patch sizes, but at the cost of higher floating-point operations (FLOPS) and memory consumption \cite{chen2021crossvit}. To relieve this problem, CrossViT \cite{chen2021crossvit} 
proposed a dual-branch ViT
where each branch operates at a different patch
size, as its own ``view'' of the image. 
The cross-attention module is then used to fuse information between the branches in order to balance the patch sizes and complexity. Similar to ViT, the final hidden vector obtained from the \texttt{CLS} tokens from the two branches are then used for image classification.

\subsection{Our Proposed RGT Model}
\label{model}

\begin{figure*}[t!]
\centering
\includegraphics[width=\linewidth]{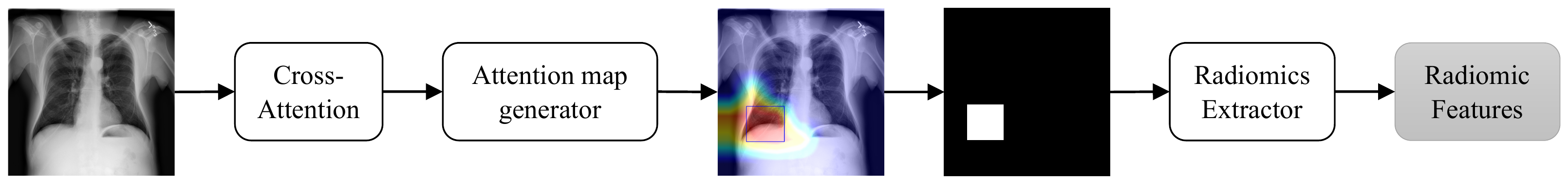}
\caption{Overview of our Bootstrap Your Own Attention (BYOA) module. For the input chest X-rays, we look at the self-attention of the \texttt{CLS} token of the Image branch on the heads of the final output of the cross attention module. Then we apply a threshold of 0.1, meaning we only keep the top 10\% of pixels in the generated attention map, to produce bounding boxes. Then with the generated bounding boxes, we use the \textit{Pyradiomics} tool to extract radiomic features from the region of interest.}
\label{byoa} 
\end{figure*}

CrossViT supplies a graceful framework to simultaneously process and fuse two different ``views'' from the same input data (e.g., different-size image patches in the original paper) \cite{chen2021crossvit}. In RGT, we extend this idea by treating the raw image itself as one ``view'' and the radiomic feature extracted from this image as another ``view'' (Fig.~\ref{rgt}). The global image representation and local radiomics representations are then fused by interacting through cross-attention. Here, a Transformer serves as the modality-agnostic backbone for both views.  

Specifically, we introduce a dual-branch cross-attention Transformer where the first (primary) branch operates on the image, while the second (auxiliary) branch handles the radiomic features. To resolve the ``chicken-and-egg'' dilemma in extracting reliable radiomic features without bounding boxes, we have designed a novel \textit{\textbf{B}ootstrap \textbf{Y}our \textbf{O}wn \textbf{A}ttention} (\textbf{BYOA}) module, using a feedback loop to learn pathology localization for radiomic feature extraction. A simple yet effective module is also utilized to fuse information between the branches. In the subsequent sections, we will describe the two branches, the BOYA module, and the fusion module.

\textbf{Image Branch.} The primary image branch uses a Progressive-Sampling ViT (PS-ViT) \cite{yue2021vision} as its backbone. 
Unlike the vanilla ViT that splits images into fixed-size tokens, PS-ViT utilizes an iterative and progressive sampling strategy to locate discriminative regions and avoid over-partitioning object structures. We experimentally observed PS-ViT outperforms ViT and other variants in our framework because it generates higher-quality and more structure-aware attention maps, which are crucial for estimating the pathology localization during training.

\textbf{Radiomics Branch.} The complementary radiomics branch is used to process and learn deep representations of radiomic features. Handcrafted features can encompass a wide range of categories, such as first-order (basic intensity and shaped-based features), second-order (texture features extracted from various matrices), and more advanced features including those calculated from Fourier and wavelet transforms. Specifically, the 107 radiomic features utilized in this work come from the following categories described below:
\begin{itemize}
    \item First-order statistics measure the distribution of pixel intensities within the region of interest. Such features include energy (the measurement of the magnitude of pixel values), entropy (the measurement of uncertainty in the image values), and max/mean/median gray level intensity. In total, we extract 18 first-order radiomic features.
    \item Shape-based features -- such as mesh surface, pixel surface, and perimeter -- describe the two-dimensional size and shape of the region of interest. While RGT can only produce rectangular bounding boxes for radiomics extraction, shape-based features can still be useful to quantify the size and aspect ratio of the extracted region. A total of 14 shape features are used in this work.
    \item Gray-level features describe statistical patterns in the pixel intensity values, drawn from the Gray Level Co-occurrence Matrix (GLCM), Gray Level Size Zone Matrix (GLSZM), Gray Level Run Length Matrix (GLRLM), Neighboring Gray Tone Difference Matrix (NGTDM), and Gray Level Dependence Matrix (GLDM). In particular, we compute 24 GLCM features, 16 GLSZM features, 16 GLRLM features, 5 NGTDM features, and 14 GLDM features.
\end{itemize}

For this branch, we use a vanilla Transformer \cite{liu2017attention} as the radiomics encoder. Please note that the only difference is that the positional encoding module is discarded, since there does not exist any positional relationship between individual radiomic features.


\textbf{Bootstrap Your Own Attention (BYOA): A Feedback Loop Module.} Our main \textit{roadblock} concerns how to generate robust radiomic features without pathology localization. On one hand, radiomic features are highly sensitive to the choice of local region of interest, for which we have no bounding box annotation. On the other hand, image features would benefit from the guidance of radiomics that encode important domain-specific quantitative features. The learning of image and radiomic features thus mutually depend on each other, forming a challenging chicken-and-egg problem.

To address this issue, we design \textbf{BYOA} to constitute an end-to-end feedback loop that can bootstrap accurate pathology localization from image data without any bounding box annotations (Fig.~\ref{byoa}). BYOA contains two components: attention map generation and radiomic feature extraction.
\begin{itemize}
    \item \textit{Attention Map Generation.} Similar to the approach in Caron \textit{et al.} \cite{caron2021emerging}, we extract self-attention of the \texttt{CLS} token from the heads of the last layer. RGT produces two \texttt{CLS} tokens from two branches, but the attention maps \textit{only} come from the image branch. To generate bounding boxes for radiomic features extraction, we first apply a threshold on the learned self-attention maps. This threshold, controlling the percentage of most responsive pixels kept for further processing, will influence the size of the resulting bounding box and thus the quality of radiomic features. After thresholding the attention map, image processing steps including a maximum filter and five consecutive binary dilations are used to ``grow'' the region of interest and smooth boundaries. Then, connected-components labeling is performed, after which we find the ``center of mass'' of each component. If this center of mass pixel is in the top decile of intensity values, a bounding box is drawn around it according to the mean height and width of the known bounding box annotations for the given disease class of interest. Here, we utilize one kind of prior knowledge of different diseases, e.g. Cardiomegaly usually occurs in the heart area, and localized Pneumonia usually occurs in the lung area, and the information of the average bounding boxes of these diseases could be seen as one kind of free-available prior knowledge, which could improve the accuracy of our model. And as one limitation of our method, for stable training, our method will generate per-class identically-sized bounding boxes. But during testing, we relaxed the setting of the bounding box generation.
    \item \textit{Radiomic Features Extraction.} Given the original images and generated bounding boxes, we used Pyradiomics \cite{van2017computational} to extract a variety of radiomic features, including 18 first-order features, 14 shape-based features, and 73 gray-level features (see Appendix for full list). For feature extraction, we adopt the default settings of PyRadiomics version 3.0.1, which includes no spatial resampling, discretization, rescaling, or normalization; this is not necessary, as input radiographs have already been min-max normalized as part of model preprocessing. All features are derived from the original image (no wavelet, Laplacian of Gaussian, or other filters are applied before feature extraction).
\end{itemize}

\textbf{Cross-Attention Fusion Module.} To aggregate global image information with local radiomics information, this fusion step involves the \texttt{CLS} token of the image branch and patch tokens of the radiomics branch, similarly, it also involves the \texttt{CLS} token of the radiomics branch and patch tokens of the image branch. As the \texttt{CLS} token is the aggregate representation of the branch, this interaction helps include information from multiple scales.
Please refer to Chen \textit{et al.} \cite{chen2021crossvit} for more details about the cross-attention mechanism.

\subsection{Semi-Supervised Loss Function}
In our framework, we aim to make the learned image features from the \texttt{CLS} token similar to the learned radiomic features in order to localize pathologies in the chest X-rays. As shown in Fig.~\ref{rgt}, RGT is trained using the linear combination of the supervised classification and unsupervised contrastive losses. For the supervised classification, considering that the chest X-ray dataset is usually highly imbalanced, we adopt the Focal Loss \cite{lin2017focal}. For unsupervised contrastive learning, we use the cross-view contrastive loss \cite{chen2020simple}. 

\textbf{Supervised Classification Focal Loss.}
We feed the output of the \texttt{CLS} tokens $I_{cls}$ (from the image branch) and $R_{cls}$ (from the radiomics branch) to a simple linear classifier. The supervised classification focal loss $\mathcal{L}_{f l}$ is defined as 
\begin{equation}
    \mathcal{L}_{f l}=\left\{\begin{array}{ll}
-\alpha\left(1-y^{\prime}\right)^{\gamma} \log y^{\prime}, & y=1 \\
-(1-\alpha) y^{\prime \gamma} \log \left(1-y^{\prime}\right), & y=0
\end{array}\right.
\end{equation}
The hyperparameter $\alpha$ allows us to give different importance to positive and negative examples, whereas $\gamma$ is used to distinguish easy and hard samples , forcing the model to place more emphasis on difficult examples. 

\textbf{Unsupervised Cross-View Contrastive Loss.} 
Our contrastive loss extends the normalized temperature scaled cross-entropy loss (NT-Xent). The difference is that we maximize agreement between two feature views extracted from different input formats, one from the image and the other from radiomic features. 

Given an anchor chest X-ray in a minibatch, the positive sample will be its radiomic feature view, and the negative samples will be other chest X-rays (both image and radiomics views). Since the \texttt{CLS} token can be regarded as the representation of the input modality, we only need to maximize the agreement between each modality's \texttt{CLS} tokens. Suppose $I_{cls, k}$ and $R_{cls, k}$ are the $k-$th image features and radiomic features in the minibatch, respectively, and $sim(\cdot)$ the cosine similarity. Then the contrastive loss function $\mathcal{L}_{c l}$ is defined as
\begin{equation}
    \mathcal{L}_{c l} = -\log\frac{\exp(sim(g_i(I_{cls, k}), g_r(R_{cls, k}))/\tau)}{\sum_{k=1 }^{N} \exp(sim(g_i(I_{cls, k}), g_r(R_{cls, k}))/\tau)}
\end{equation}
where $\tau$ is the temperature. The final contrastive loss is summed over all instances in the minibatch. 

Overall, we treat RGT training as a weakly-supervised multi-task learning problem. In our chest X-ray setting, there exist two types of labels: disease class labels and pathology bounding box annotations. In our case, we \textit{only} use the disease labels for training, even though the ultimate goal is to accurately localize those pathologies. Here, when we say ``weakly-supervised'' localization, we mean that we are able to localize pathologies only using supervision from whole-image disease labels. The combined loss function for supervised disease classification and unsupervised cross-view contrastive learning is as follows:
\begin{equation}
    \mathcal{L}= (1 -\lambda) \times \mathcal{L}_{cl}+ \lambda \times \mathcal{L}_{fl}
\end{equation}

\section{Experiments}
\subsection{Dataset and Protocol Setting} 
The NIH ChestXRay dataset \cite{wang2017chestx} consists of 112,120 chest X-rays collected from 30,805 patients, where each image is labeled with one or more of 14 cardiopulmonary diseases.
The labels are extracted from the associated radiology report using an automatic labeler \cite{peng2018negbio} with a reported accuracy of 90\%.
For a subset of 880 images, the NIH dataset also provides bounding box localizations associated with eight disease classes: Atelectasis, Cardiomegaly, Effusion, Infiltration, Mass, Nodule, Pneumonia, and Pneumothorax. The remaining six diseases are diffuse in nature, meaning it is not clinically meaningful to provide a ``localization'' for these pathologies. Since this study aims to develop a model for weakly supervised disease localization, we only proceed with the eight diseases which have ground truth bounding box annotations. Specifically, we only use the image-level disease labels for these eight focal diseases to train RGT, binning all other classes into the already provided ``No Findings'' category.
A significant difference between our method and existing baseline methods for pathology localization \cite{Liu_2019_ICCV, li2018thoracic} is that our method does not require any training data related to the bounding box while others use some percentage of these images for training.

In our experiments, we followed the same protocol as in related studies \cite{wang2017chestx, li2018thoracic}, randomly partitioning the dataset (excluding images with bounding box annotations) into three subsets: 70\% for training, 10\% for validation, and 20\% for testing. In order to prevent data leakage across patients, we make sure that there is no patient overlap between our train, validation, and test set.


\subsection{Implementation Details}
We build our image branch encoder based on PS-ViT \cite{yue2021vision}, and apply their default hyperparameters for training.
We use a shallower image encoder than the original PS-ViT, using 6 layers.
For the radiomic branch encoder, since the radiomic features are already informative features, we use a small standard Transformer (2 layers) to learn representations of the radiomic features. We then add one more cross-attention layer to fuse the learned image features with the learned radiomic features. We set the batch size to 128 and train the model for 50 epochs. We used a cosine linear-rate scheduler with a linear warm-up of 5 epochs, an initial learning rate of 0.004, and a weight decay of 0.05. We downscale the images to 224\texttimes224 and normalize based on the mean and standard deviation of images in the ImageNet training set. We also augment the training data with random horizontal flipping. During the evaluation, we resize the image to 256\texttimes256 and take the center crop 224\texttimes224 as the input.

\begin{table*}[!t]
\centering
\caption{Weakly supervised pathology localization results on the NIH ChestXRay dataset as measured by IoU accuracy at a fixed threshold. Please note that since RGT was solely supervised by disease class labels (not pathology localizations), we only compare localization performance with previous methods following the same setting for fair evaluation.}
\resizebox{1\textwidth}{!}{
\setlength{\tabcolsep}{4pt}
\begin{tabular}{clccccccccc}
\toprule
\textbf{T(IoU)} & {Model} &{Atelectasis} & {Cardiomegaly} & {Effusion} & {Infiltration} & {Mass} & {Nodule} & {Pneumonia} & {Pneumothorax} & \textbf{Mean}\\
\midrule
\multirow{3}{*}{0.1}
& Wang \textit{et al.} \cite{wang2017chestx} & 0.69 & 0.94 & 0.66 & 0.71 & 0.40 & 0.14 & 0.63 & 0.38 & 0.569\\
&ViT & 0.58 & 0.91 & 0.61 & 0.77 & 0.44 & 0.11 & 0.75 & 0.25 & 0.553 \\
&RGT & 0.61 & 0.95 & 0.65 & 0.82 & 0.50 & 0.13 & 0.79 & 0.28 & \textbf{0.591} \\
\midrule
\multirow{3}{*}{0.2}
& Wang \textit{et al.} \cite{wang2017chestx} & 0.47 &0.68 &0.45& 0.48& 0.26 &0.05 &0.35 &0.23& 0.371\\
&ViT & 0.38 & 0.85 & 0.39 & 0.55 & 0.24 & 0.01 & 0.51 & 0.15 & 0.385 \\
&RGT & 0.41 & 0.91 & 0.41 & 0.59 & 0.26 & 0.05 & 0.57 & 0.19 & \textbf{0.424} \\
\midrule
\multirow{3}{*}{0.3}
& Wang \textit{et al.} \cite{wang2017chestx} & 0.24 &0.46 &0.30& 0.28& 0.15& 0.04& 0.17 &0.13 &0.221\\
&ViT & 0.20 & 0.45 & 0.19 & 0.32 & 0.06 & 0.00 & 0.21 & 0.02 & 0.181 \\
&RGT & 0.28 & 0.79 & 0.22 & 0.38 & 0.12 & 0.01 & 0.41 & 0.05 & \textbf{0.283} \\
\midrule
\multirow{3}{*}{0.4}
& Wang \textit{et al.} \cite{wang2017chestx} & 0.09& 0.28& 0.20& 0.12& 0.07& 0.01& 0.08& 0.07& 0.115\\
&ViT & 0.10 & 0.21 & 0.03 & 0.05 & 0.02 & 0.00 & 0.04 & 0.00 & 0.056 \\
&RGT & 0.17 & 0.54 & 0.13 & 0.18 & 0.07 & 0.01 & 0.26 & 0.02 & \textbf{0.173}\\
\midrule
\multirow{3}{*}{0.5}
& Wang \textit{et al.} \cite{wang2017chestx} & 0.05 &0.18& 0.11& 0.07& 0.01& 0.01& 0.03& 0.03& 0.061\\
&ViT & 0.05 & 0.15 & 0.01 & 0.04 & 0.02 & 0.00 & 0.03 & 0.00 & 0.034 \\
&RGT & 0.08 & 0.32 & 0.05 & 0.09 & 0.05 & 0.00 & 0.12 & 0.01 & \textbf{0.090} \\
\midrule
\multirow{3}{*}{0.6}
& Wang \textit{et al.} \cite{wang2017chestx} & 0.02 &0.08 &0.05 &0.02 &0.00 &0.01 &0.02 &0.03 &0.029\\
&ViT & 0.01 & 0.03 & 0.01 & 0.01 & 0.01 & 0.00 & 0.01 & 0.00 & 0.010 \\
&RGT & 0.02 & 0.15 & 0.03 & 0.04 & 0.03 & 0.00 & 0.06 & 0.00 & \textbf{0.041} \\
\midrule
\multirow{3}{*}{0.7}
& Wang \textit{et al.} \cite{wang2017chestx} & 0.01 &0.03 &0.02 &0.00& 0.00 &0.00 &0.01& 0.02& 0.011\\
&ViT & 0.00 & 0.00 & 0.00 & 0.01 & 0.00 & 0.00 & 0.00 & 0.00 & 0.001 \\
&RGT & 0.01 & 0.04 & 0.01 & 0.02 & 0.01 & 0.00 & 0.03 & 0.00 & \textbf{0.015} \\
\bottomrule
\end{tabular}}
\label{tab:localizationresult}
\end{table*}

\begin{figure*}[!t]
\centering
\includegraphics[width=\linewidth]{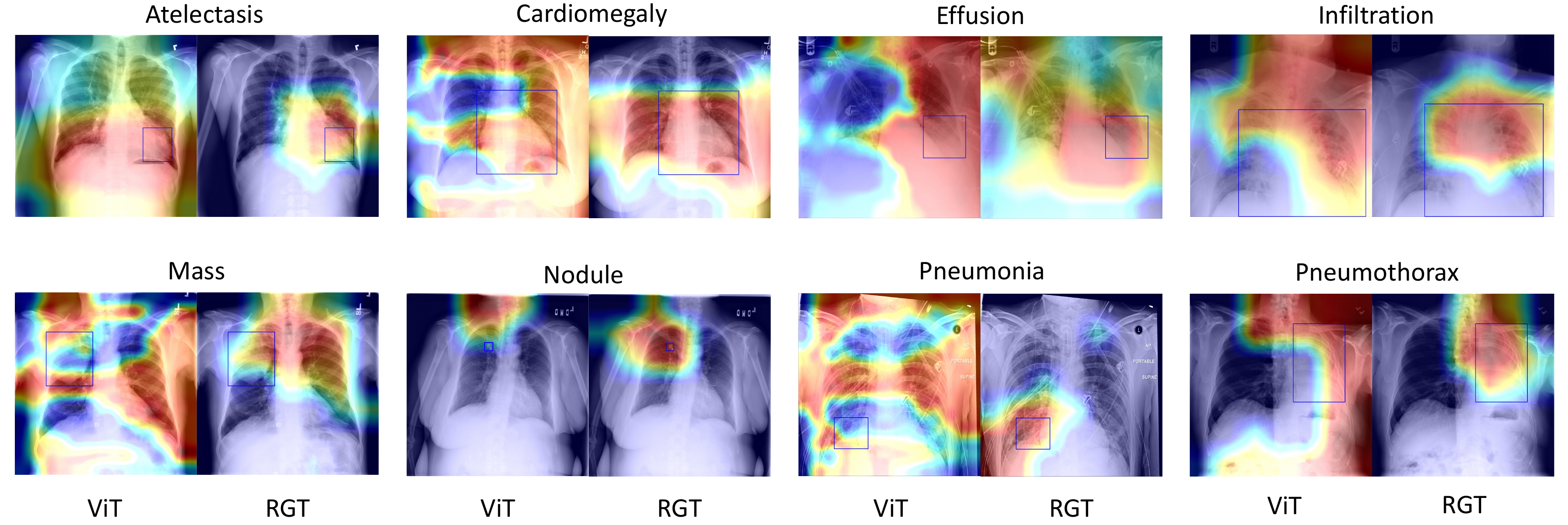}
\caption{Example visualizations of pathology localization when evaluated on the 880 NIH ChestXRay images with bounding box annotations. The attention maps are generated from the self-attention maps of the \texttt{CLS} token. The ground-truth bounding boxes are shown in blue. The left image in each pair is the localization result of ViT \cite{dosovitskiy2020image}, and the right one is our localization results obtained by RGT. All examples are positive for the corresponding disease labels. Best viewed in color.}
\label{attention}
\end{figure*}

\subsection{Pathology Localization}
The NIH Chest X-ray dataset contains 880 images labeled by radiologists with bounding box information, which we use to evaluate the performance of RGT for pathology localization. Many prior works \cite{li2018thoracic, Liu_2019_ICCV} have used a fraction of ground truth (GT) bounding boxes for training and evaluated their system on the remaining examples. Unlike these approaches, RGT \textit{uses no bounding box annotations during training}, only using the subset of bounding box-annotated images for evaluation. Table \ref{tab:localizationresult} presents our evaluation results on all 880 images. We used \cite{wang2017chestx} as our baseline to compare our localization results since it follows the same experimental setting of weakly supervised training on only disease labels.

\begin{table*}[!t]
\centering
\caption{Pathology classification results for CNN- and Transformer-based methods on the NIH ChestXRay dataset, as measured by AUC. For each column, \textbf{bold} values denote the best results for the given disease class. For RGT, the average AUC per class is presented, with the standard deviation in parentheses, across three training runs with different random initializations.}
\setlength{\tabcolsep}{4pt}
\resizebox{1\textwidth}{!}{%
\begin{tabular}{@{}l@{}ccccccccc@{}}
\toprule
\multicolumn{1}{c}{Method} & Atelectasis  & Cardiomegaly  & Effusion  & Infiltration  & Mass  & Nodule  & Pneumonia  & Pneumothorax  & \textbf{Mean}\\ 
\midrule
CNN\\
\hspace*{1em} Wang \textit{et al.} \cite{wang2017chestx} &  0.72 & 0.81 & 0.78 & 0.61 & 0.71 & 0.67 & 0.63 & 0.81 & 0.718\\
\hspace*{1em} Wang \textit{et al.} \cite{wang2018tienet} &  0.73  & 0.84  & 0.79    & 0.67  & 0.73  &  0.69 &  0.72  &  0.85 & 0.753\\
\hspace*{1em} Yao \textit{et al.} \cite{yao2017learning} &  0.77 & 0.90 & 0.86 & 0.70 & 0.79 & 0.72 & 0.71 & 0.84 & 0.786\\
\hspace*{1em} Rajpurkar \textit{et al.} \cite{rajpurkar2017chexnet} & 0.82 & 0.91 & \textbf{0.88} & 0.72 & 0.86 & 0.78 & 0.76 & \textbf{0.89} & 0.828\\
\hspace*{1em} Kumar \textit{et al.} \cite{kumar2018boosted} & 0.76 & 0.91 & 0.86 & 0.69 & 0.75 & 0.67 & 0.72 & 0.86 & 0.778\\ 
\hspace*{1em} Liu \textit{et al.} \cite{liu2019align} & 0.79 & 0.87 & \textbf{0.88} & 0.69 & 0.81 & 0.73 & 0.75 & \textbf{0.89} & 0.801\\
\hspace*{1em} Seyyed \textit{et al.} \cite{seyyed2020chexclusion} & 0.81 & \textbf{0.92} & 0.87 & 0.72 & 0.83 & 0.78 & 0.76 & 0.88 & 0.821\\ 
\hspace*{1em} Han \textit{et al.} \cite{han2020using} & \textbf{0.83} & \textbf{0.92} & 0.87 & 0.76 & 0.85 & 0.76 & 0.77 & 0.86 & 0.828\\ \midrule
Transformer\\
\hspace*{1em}ViT & 0.74 & 0.78 & 0.81 & 0.72 & 0.70 & 0.66 & 0.65 & 0.76 & 0.728\\
\hspace*{1em}CrossViT & 0.69 & 0.71 & 0.72 & 0.72 & 0.74 & 0.79 & \textbf{0.82} & 0.88 & 0.759 \\
\hspace*{1em}PS-ViT & 0.75 & 0.81 & 0.82 & 0.73 & 0.79 & 0.73 & 0.69 & 0.81 & 0.766\\
\hspace*{1em}RGT (ours) & 0.80  & \textbf{0.92} & 0.78 & \textbf{0.86} & \textbf{0.88} & \textbf{0.88} & 0.79 & 0.81 & \textbf{0.839}\\
 & ($\pm 0.02$) & ($\pm 0.00$) & ($\pm 0.01$) & ($\pm 0.01$) & ($\pm 0.02$) & ($\pm 0.00$) & ($\pm 0.01$) & ($\pm 0.02$) & --\\
\bottomrule
\end{tabular}}

\label{AUC}
\end{table*}

\subsubsection{Evaluation Metric} For localization, we evaluated our detected regular rectangular regions against the annotated bounding boxes, using a thresholded \textbf{IoU accuracy}, following Wang \textit{et al.} \cite{wang2017chestx}. Our localization results are only calculated for the 880 images that have ground truth annotation for 8 diseases. To compute IoU accuracy, the localization is defined as ``correct'' only if the observed IoU between the predicted and ground truth localization exceeds a fixed IoU threshold, T(IoU). We evaluated RGT for different thresholds ranging from \{0.1, 0.2, 0.3, 0.4, 0.5, 0.6, 0.7\} as shown in Table \ref{tab:localizationresult}.

\subsubsection{Comparison with Prior Works} We compared disease localization accuracy under varying IoU with baselines following the same training setting as RGT (Table \ref{tab:localizationresult}). Unlike other baselines\cite{li2018thoracic, Liu_2019_ICCV} that use a portion of 880 images for evaluation (because they need the remaining data for training), we used all 880 annotated images for evaluation. Therefore, no $k$-fold cross-validation for localization was performed.
RGT average localization performance across 8 diseases is considerably higher than the baseline under all IoU thresholds. When the IoU threshold is set to 0.1, RGT outperforms the baseline \cite{wang2017chestx} in the Cardiomegaly, Infiltration, Mass, and Pneumonia classes. 
Even with higher thresholds, our model is superior to the baseline. For example, when evaluated at T(IoU) = 0.5, our ``Cardiomegaly'' accuracy is 32\%, while the reference model achieves 18\%. Similarly, our ``Pneumonia'' accuracy is 12\%, while the reference model reaches 3\% accuracy. 
Note that some diseases can appear in multiple locations, but the ground truth might have mentioned only one such location. This can significantly impact the accuracy at high thresholds. 

\subsubsection{Discussion of Visualization} More importantly, we also include our own trained ViT as an additional baseline here. The quantitative results above demonstrate that, compared to the standard ViT, the additional radiomics branch and BYOA module enable RGT to learn more accurate and fine-grained pathology localizations. Example visualizations of localization results of both ViT and RGT can be seen in Fig.~\ref{attention}.
We can observe that RGT produces qualtitatively more accurate localizations than ViT for all diseases, but particularly Atelectasis, Cardiomegaly, Infiltration, Pneumonia, and Pneumothorax. Visualizations for most diseases reveal that both models often attend to regions outside the clinically relevant region of interest. However, RGT consistently attends to a smaller number of ``extraneous'' pixels than the standard ViT. Further, RGT always contains a significant portion of the ground truth localized region, while the ViT attention map does not -- for example, see Nodule, Pneumonia, and Pneumothorax.


\subsection{Pathology Classification}
Pathology classification for chest X-rays is a multi-label classification problem. The objective is to assign one or more labels (among 8 cardiopulmonary diseases: Atelectasis, Cardiomegaly, Effusion, Infiltration, Mass, Nodule, Pneumonia, and Pneumothorax) to each input image at inference time. We compared RGT with related reference approaches, which represent state-of-the-art disease classifcation performance on the NIH ChestXRay dataset. For RGT, we report the average AUC of 3 runs to show the robustness of our model.

\subsubsection{Evaluation Metric}
We used Area under the Receiver Operating Characteristic Curve (AUC) to estimate the performance of our model \cite{FawcettTom2006AnIT}. A higher AUC score implies a model that is more capable of discriminating between classes. We also provide mean AUC across all the classes to highlight the overall performance of our model.

\subsubsection{Comparison with Prior Works}
AUC scores for each disease and mean AUC across eight diseases are presented in Table \ref{AUC}. We not only compared RGT with previous CNN-based state-of-the-art (SOTA) models, but also several Transformer-based models.
We find that RGT outperformed all baseline approaches with respect to mean AUC across all diseases; specifically, RGT reached 0.839 mean AUC, outperforming the previous SOTA for disease classification \cite{rajpurkar2017chexnet} by a margin of 0.011. When considering classification performance on individual disease classes, RGT also achieved best performance on four of the eight classes. Our proposed model outperformed the next-best baseline by a margin 0.13 AUC for Infiltration, 0.09 for Nodule, and 0.02 for Mass.
Compared to the Transformer-based models, the key difference is that we utilize the extracted radiomic features for disease prediction, improving the classification accuracy and enriching the model's interpretability due to the utilization of handcrafted radiomic features. Please note that Liu \textit{et al.} \cite{liu2019align} used 5-fold cross-validation in their model evaluation. While the problem settings are very similar, the evaluation schemes are so different that a direct comparison of this work to RGT would be inappropriate.

\begin{figure}[htbp]
    \centering
    \includegraphics[width=\linewidth]{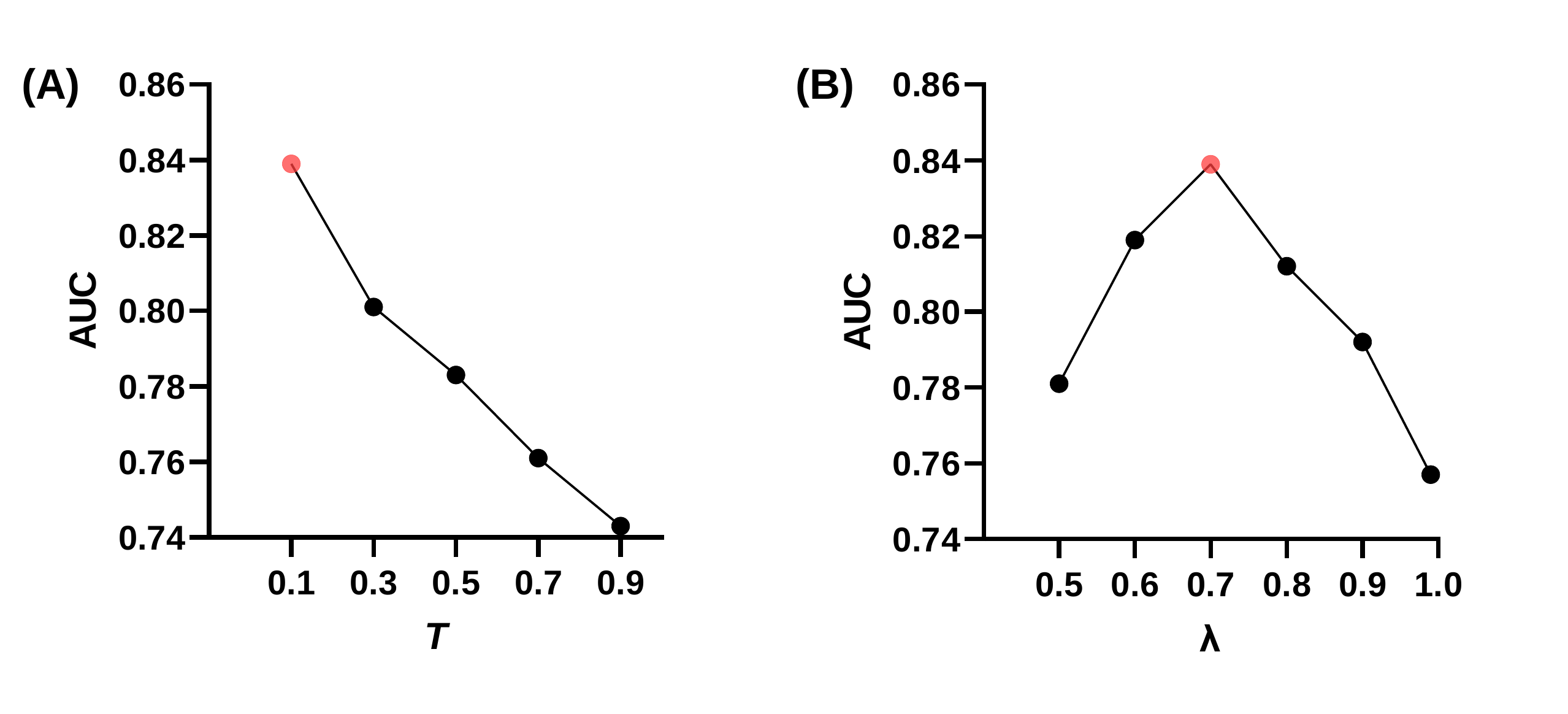}
    \caption{Effect of (A) varying $T$ in attention map generation and (B) varying $\lambda$ in Equation (3) on pathology classification for the NIH ChestXRay dataset.}
    \label{fig:auc}
\end{figure}

\subsubsection{Effect of Attention Map Threshold}
\label{thresholds}
We investigate the impact of the threshold $(T)$, used in the process of attention map generation, on the performance of RGT for disease classification. Fig.~\ref{fig:auc}A summarizes the AUC comparison of RGT for different values of $T$. Higher values of $T$ imply smaller bounding boxes from which to extract radiomic features.
During our experiments, we found that RGT performs better on the disease classification task when larger bounding boxes are generated. Since radiomic features are typically computed for highly localized -- often small -- regions of interest, this was originally an unintuitive finding. There appears to be a tradeoff between bounding box size and the resulting performance on the disease classification and localization tasks. Specifically, extracting smaller boxes that are accurately localized and ignore as much background signal as possible should lead to more robust and useful radiomic features. However, attending to smaller regions of the image comes at the expense of decreasing the ``receptive field'' of learned global image features, thus degrading the quality of the classification task. This observation emphasizes the difference between disease classification and localization tasks: global information aids classification while rich local information aids localization.

\subsubsection{Effect of Contrastive Learning}

We also investigated the impact of the unsupervised contrastive loss on RGT's disease classification ability. Specifically, we evaluate the performance of RGT for disease classification by varying $\lambda$ in equation (3). Fig.~\ref{fig:auc}B summarizes the AUC comparison of RGT for different values of $\lambda$. Higher values of $\lambda$ implies lower weight to contrastive loss. During our experiments, we found that RGT performs worse when small weight (1\%) is given to the contrastive loss. RGT's performance improves when we increase contrastive loss weight, but after a certain point ($\lambda = 0.7$), performance considerably decreases. This confirms our hypothesis that both contrastive and focal losses are important, but that care must be taken to properly balance the objectives. The supervised classification and unsupervised constrastive losses enable RGT to learn both disease-level and patient-level discriminative visual features.

\begin{figure*}[!t]
\centering
\includegraphics[width=\linewidth]{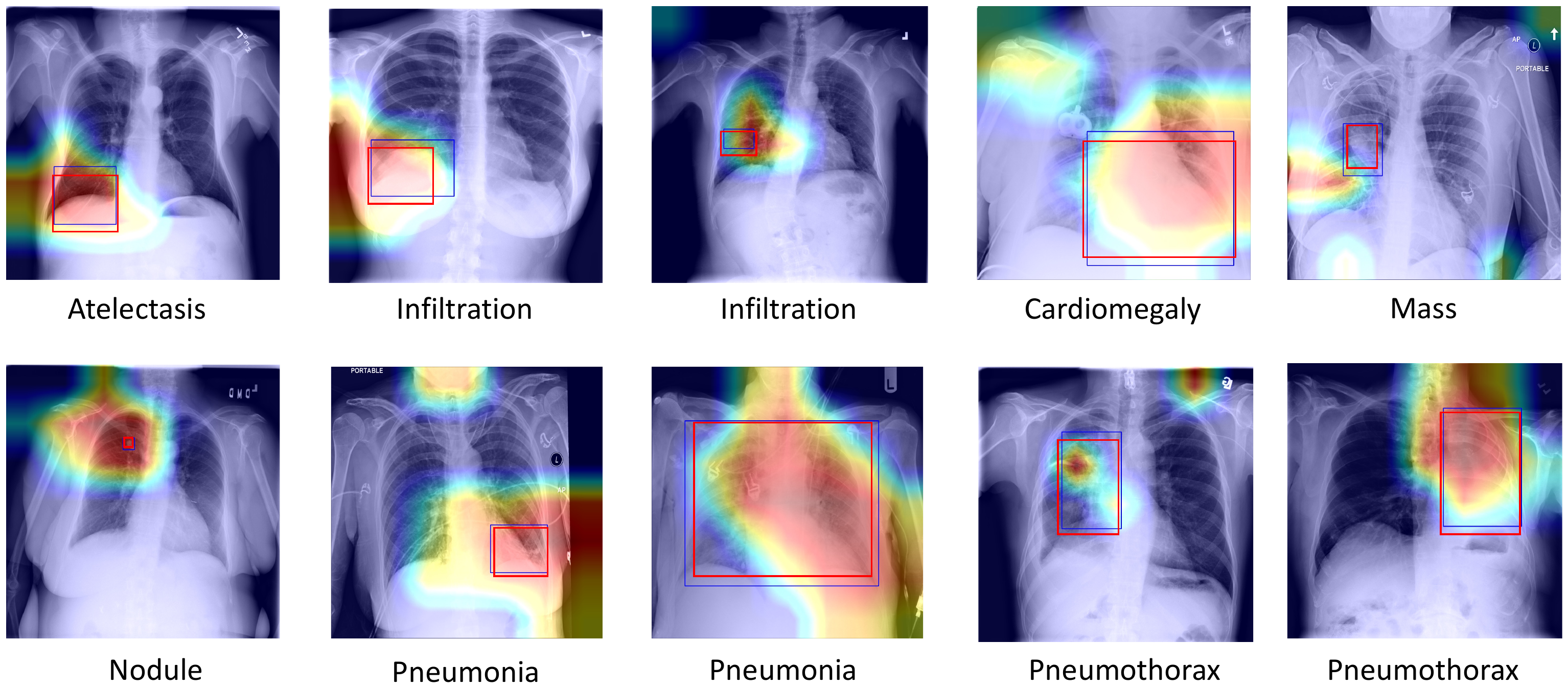}
\caption{System Usability Study. Visualizations of pathology localization come from 10 randomly selected chest X-rays from the NIH ChestXRay dataset that do not have ground truth localization annotations. Saliency heat maps are generated from the self-attention maps of the \texttt{CLS} token from our trained RGT model. The red and blue represent the pathology localizations provided by two radiologists, who were instructed to draw a rectangular bounding box around the most salient image region in 90 seconds.}
\label{sus}
\end{figure*}







\subsection{System Usability Study}
We hired two radiologist experts to validate the usefulness of RGT’s disease localizations; one expert has 3.5 years of experience, while the other has 5. For this additional study, we randomly selected 10 images from the NIH ChestXRay dataset that did not have ground truth bounding box annotations. We then used RGT to predict the disease classification and localization visualization. Finally, we asked two radiologists to provide their own pathology localization for each image. Each radiologist was instructed to draw a rectangular bounding box around the clinically relevant region of interest within 90 seconds. Results can be seen in Fig.~\ref{sus}. Each image contains two human-annotated bounding boxes (red is Expert 1, and blue is Expert 2) and the extracted attention map from our RGT.

For disease classification, the two radiologists agreed with RGT’s prediction for all ten cases. For the localization task, we observe that the inter-rater consistency between two radiologists is very high, suggesting that they clearly agreed on the most salient image region. Overall, the radiologists found the RGT attention maps to significantly overlap with their own localizations, demonstrating the usefulness of our approach. With the exception of the “Mass” example, there is a strong agreement between the most responsive pixels in RGT’s heat map and the radiologists’ annotations.

\subsection{Limitations and Discussion}
There exist two main limitations to this approach. One limitation is the fact that self-attention provides only a coarse approximation of salient regions unless trained on extremely large amounts of data (e.g., see DINO \cite{caron2021emerging}). Without this scale of chest radiography data available, other principled methods for saliency visualization may provide more fine-grained localizations for radiomics extraction than the native self-attention of our proposed RGT architecture. For example, the Anchors approach of Ribeiro \textit{et al.} \cite{ribeiro2018anchors} or other input space visualization methods like LIME \cite{ribeiro2016should}, SHAP \cite{lundberg2017unified}, and deep Taylor decomposition \cite{montavon2017explaining} could be used in place of our proposed heatmap generation process. Future work will consider adapting such approaches to generate more accurate localizations for improved radiomics feature extraction, and thus better downstream disease classification and localization.


Another limitation of this approach is the fact that fixed-sized bounding boxes per target disease class are generated during RGT training. This would, for example, make it difficult to distinguish the visual presentation of diffuse vs. localized pneumonia. However, this can be alleviated with finer granularity in the image-level disease labels used to train RGT; for instance, if ``diffuse'' and ``localized'' pneumonia were distinct class labels, then RGT would be able to provide visually distinct localizations of the two conditions. Future work may involve a module that learns the optimal bounding box dimensions for each disease in an unsupervised manner. Alternatively, the adoption of other saliency visualization methods instead of RGT's self-attention -- as explained in the previous paragraph -- may resolve this limitation of fixed-size bounding boxes per disease class.

\section{Conclusion}
In this paper, we propose a radiomics-guided cross-attention Transformer, RGT, that can jointly localize and classify abnormalities in chest X-rays without supervision from bounding box annotations. Our approach differs from previous related studies in the choice of a unified Transformer architecture, the use of radiomic features, and a feedback loop for image and radiomic features to mutually interact during the training process. This work aims to bring the field of computer-aided diagnosis closer to clinical practice by making domain-specific quantitative features (in the form of radiomics) more accessible to automated medical image analysis tools, with the hope that this will increase the model's interpretability. Experimental results demonstrate that our method outperforms state-of-the-art algorithms in this weakly supervised setting, particularly for disease localization, where our method can generate more accurate and clinically useful bounding boxes.

\bibliography{ref}
\bibliographystyle{IEEEtran}

\begin{table*}[!ht]
\centering
\caption{Name and type of all 107 radiomic features used in this study. Feature names are defined by PyRadiomics, and detailed descriptions of each feature can be found in the PyRadiomics documentation: \url{https://pyradiomics.readthedocs.io/en/latest/features.html}. GLCM = gray level co-occurrence matrix; GLDM = gray level dependence matrix, GLRLM = gray level run-length matrix; GLSZM = gray level size zone matrix; NGTDM = neighboring gray tone distance matrix.}
\begin{tabular}{@{}lclc@{}}
\toprule
\multicolumn{1}{c}{PyRadiomics Feature} & \multicolumn{1}{c}{Feature Type} & \multicolumn{1}{c}{PyRadiomics Feature} & \multicolumn{1}{c}{Feature Type} \\
\midrule
Elongation & Shape & SumEntropy & GLCM \\
Flatness & Shape & SumSquares & GLCM \\
LeastAxisLength & Shape & DependenceEntropy & GLDM \\
MajorAxisLength & Shape & DependenceNonUniformity & GLDM \\
Maximum2DDiameterColumn & Shape & DependenceNonUniformityNormalized & GLDM \\
Maximum2DDiameterRow & Shape & DependenceVariance & GLDM \\
Maximum2DDiameterSlice & Shape & GrayLevelNonUniformity & GLDM \\
Maximum3DDiameter & Shape & GrayLevelVariance & GLDM \\
MeshVolume & Shape & HighGrayLevelEmphasis & GLDM \\
MinorAxisLength & Shape & LargeDependenceEmphasis & GLDM \\
Sphericity & Shape & LargeDependenceHighGrayLevelEmphasis & GLDM \\
SurfaceArea & Shape & LargeDependenceLowGrayLevelEmphasis & GLDM \\
SurfaceVolumeRatio & Shape & LowGrayLevelEmphasis & GLDM \\
VoxelVolume & Shape & SmallDependenceEmphasis & GLDM \\
10Percentile & First-Order & SmallDependenceHighGrayLevelEmphasis & GLDM \\
90Percentile & First-Order & SmallDependenceLowGrayLevelEmphasis & GLDM \\
Energy & First-Order & GrayLevelNonUniformity & GLRLM \\
Entropy & First-Order & GrayLevelNonUniformityNormalized & GLRLM \\
InterquartileRange & First-Order & GrayLevelVariance & GLRLM \\
Kurtosis & First-Order & HighGrayLevelRunEmphasis & GLRLM \\
Maximum & First-Order & LongRunEmphasis & GLRLM \\
MeanAbsoluteDeviation & First-Order & LongRunHighGrayLevelEmphasis & GLRLM \\
Mean & First-Order & LongRunLowGrayLevelEmphasis & GLRLM \\
Median & First-Order & LowGrayLevelRunEmphasis & GLRLM \\
Minimum & First-Order & RunEntropy & GLRLM \\
Range & First-Order & RunLengthNonUniformity & GLRLM \\
RobustMeanAbsoluteDeviation & First-Order & RunLengthNonUniformityNormalized & GLRLM \\
RootMeanSquared & First-Order & RunPercentage & GLRLM \\
Skewness & First-Order & RunVariance & GLRLM \\
TotalEnergy & First-Order & ShortRunEmphasis & GLRLM \\
Uniformity & First-Order & ShortRunHighGrayLevelEmphasis & GLRLM \\
Variance & First-Order & ShortRunLowGrayLevelEmphasis & GLRLM \\
Autocorrelation & GLCM & GrayLevelNonUniformity & GLSZM \\
ClusterProminence & GLCM & GrayLevelNonUniformityNormalized & GLSZM \\
ClusterShade & GLCM & GrayLevelVariance & GLSZM \\
ClusterTendency & GLCM & HighGrayLevelZoneEmphasis & GLSZM \\
Contrast & GLCM & LargeAreaEmphasis & GLSZM \\
Correlation & GLCM & LargeAreaHighGrayLevelEmphasis & GLSZM \\
DifferenceAverage & GLCM & LargeAreaLowGrayLevelEmphasis & GLSZM \\
DifferenceEntropy & GLCM & LowGrayLevelZoneEmphasis & GLSZM \\
DifferenceVariance & GLCM & SizeZoneNonUniformity & GLSZM \\
Id & GLCM & SizeZoneNonUniformityNormalized & GLSZM \\
Idm & GLCM & SmallAreaEmphasis & GLSZM \\
Idmn & GLCM & SmallAreaHighGrayLevelEmphasis & GLSZM \\
Idn & GLCM & SmallAreaLowGrayLevelEmphasis & GLSZM \\
Imc1 & GLCM & ZoneEntropy & GLSZM \\
Imc2 & GLCM & ZonePercentage & GLSZM \\
InverseVariance & GLCM & ZoneVariance & GLSZM \\
JointAverage & GLCM & Busyness & NGTDM \\
JointEnergy & GLCM & Coarseness & NGTDM \\
JointEntropy & GLCM & Complexity & NGTDM \\
MCC & GLCM & Contrast & NGTDM \\
MaximumProbability & GLCM & Strength & NGTDM \\
SumAverage & GLCM & & \\ \bottomrule
\end{tabular}
\end{table*}

\appendix

\end{document}